\documentclass{article}

\usepackage[preprint]{neurips_2026}

\usepackage[utf8]{inputenc} 
\usepackage[T1]{fontenc}    
\usepackage{hyperref}       
\usepackage{url}            
\usepackage{booktabs}       
\usepackage{amsfonts}       
\usepackage{nicefrac}       
\usepackage{microtype}      
\usepackage{xcolor}         
\usepackage{titletoc}
\usepackage{etoc}
\usepackage{enumitem}
\usepackage{amsmath}
\usepackage{tikz}
\usepackage{amssymb}

\usepackage{multirow}       

\usepackage{longtable}
\usepackage{booktabs}
\usepackage{subcaption}
\usepackage{bm}

\newcommand{\MM}{\mathcal{M}}

\title{Empirical Auditing of Edge-Private Graph Generators}

\author{%
  Anum Fatima \\
  Department of Statistics\\
  University of Oxford\\
  Oxford, United Kingdom \\
  \texttt{fatima@stats.ox.ac.uk} \\
   \And
   Stratis Limnios \\
   Plenitude Consulting, \\London, United Kingdom \\
\texttt{stratis.limnios@plenitudeconsulting.com} \\
   \AND
   James Adams \\
  National Physical Laboratory \\
  Teddington, United Kingdom \\
\texttt{james.adams@npl.co.uk} \\
   \And
   Lukasz Szpruch \\
  School of Mathematics,  \\
  University of Edinburgh, \\ Edinburgh, United Kingdom \\
  \texttt{l.szpruch@ed.ac.uk} \\
   \And
   Carsten Maple \\
   WMG, University of Warwick, \\ Coventry, United Kingdom \\
  \texttt{CM@warwick.ac.uk} \\
  \And 
  Gesine Reinert \\
  Department of Statistics\\
  University of Oxford\\
  Oxford, United Kingdom \\
  \texttt{reinert@stats.ox.ac.uk} \\
  \And
  Andrew Elliott \\
  School of Mathematics \& Statistics \\
  University of Glasgow, \\ Glasgow, United Kingdom \\ \texttt{Andrew.Elliott@glasgow.ac.uk}
}

\begin{document}

\maketitle

\begin{abstract}
We empirically audit privacy leakage by testing whether outputs from edge-neighbouring inputs remain distinguishable, using statistically valid lower bounds on the privacy loss witnessed by our attacks. Our framework compares direct-edge, local-structural, and GNN-based attacks through the geometry surrounding a target edge. Experiments across two generators and two networks show that privacy leakage is both mechanism- and network-dependent, with learned representations revealing information not captured by conventional local statistics. 
  
\end{abstract}

\section{Introduction and Motivation}

Networks (or graphs, used here interchangeably) provide a natural representation of complex relational data and are widely used in domains such as transportation \citep{transp1,transp2}, biology \citep{bio1,bio2}, and the social sciences \citep{wasserman1994social}. Based on this representation, network analysis 
can produce insights into the system in question, e.g. through derived statistics. However, if these networks encode sensitive relational information, then releasing the derived statistics may reveal said information.

One solution, differential privacy (DP), provides a principled framework for protecting sensitive information through a quantifiable probabilistic privacy guarantee. Informally, DP limits the influence of a single record on the distribution over released outputs. In graph settings, this has motivated private synthetic graph generators that aim to preserve structural properties while providing node- or edge-level DP guarantees. Despite a growing range of such mechanisms, there is no widely adopted framework for empirically auditing their claimed privacy guarantees. Existing benchmarking efforts, such as \cite{liu2024pgb}, primarily evaluate 
private graph generators' utility, while related work has used membership and link-inference attacks to study privacy leakage in DP graph learning, \cite{he2021,Guan2025}. Our work instead audits the privacy leakage of the \emph{graph-generation mechanism} itself: We propose an empirical framework for auditing differentially private graph generation mechanisms. We formulate privacy auditing as an adversarial distinguishability problem: given privatised outputs generated from edge-neighbouring graphs, we investigate whether the differing edge leaves an observable and exploitable footprint in the local structural geometry. We use attacks of increasing expressiveness to quantify the privacy loss witnessed in practice.

Our main contributions are:
\begin{itemize}[noitemsep,nosep,labelindent=\parindent,leftmargin=*, topsep=0pt]
    \item We introduce a mechanism-agnostic auditing framework for empirically auditing edge-level privacy leakage 
    in differentially private graph generators.

    \item We develop attacks that exploit direct edge information, hand-crafted local structural geometry and GNN-based neighbourhood representations to distinguish outputs generated from edge-neighbouring graphs.

    \item We derive a confidence-valid lower bound on the privacy loss witnessed by these attacks and compare it with the privacy budget claimed by the mechanism.
\end{itemize}

\section{Background and Auditing Framework}
\label{sec:background}

As in \cite{li2023private}, we call two graphs, $G=(V, E)$ and $G'=(V', E')$, \textbf{edge-neighbouring graphs}, denoted by $G\sim G'$, if they have the same node set and differ in exactly one edge, that is, $V = V'$ and
$|(E \cup E') \backslash (E \cap E')| = 1.$ A randomised graph mechanism $\mathcal{M}$ with output space $\mathcal{G}$ satisfies (pure) \textbf{${\bm{\epsilon}}$-edge differential privacy} if, for every pair of edge-neighbouring graphs $G\sim G'$ and every measurable set
$S\subseteq\mathcal{G}$,
\[\mathbb{P}[\mathcal{M}(G) \in S] \le e^{\epsilon} \mathbb{P}[\mathcal{M}(G') \in S],\]
with $P_{\epsilon}(\cdot\mid G)$ the distribution over privatised graphs induced by $\mathcal{M}$ on $G$. This differential privacy notion can be expressed as a testing problem, as in \cite{wasserman_statistical_2009}, see Appendix \ref{sec:hypdp}. For edge-neighbouring graphs \(G \sim G'\) and a (measurable) rejection region \(A \subseteq \mathcal{G}\), denote the true positive rate by 
$\operatorname{TPR}(A) = \Pr\!\left[\mathcal{M}(G) \in A\right],$ and the false positive rate by $
    \operatorname{FPR}(A)=\Pr\!\left[\mathcal{M}(G') \in A\right]
$.
As in Eq.~(4) in \cite{nasr2021adversary}, we have as a witnessed lower bound
on the privacy loss
\begin{equation} \label{eq:epswitt}
\varepsilon_{\mathrm{witness}}(A) =
\max\Big\{\, 
0,\,
\log\frac{\operatorname{TPR}(A)}{\operatorname{FPR}(A)},\,
\log\frac{1-\operatorname{FPR}(A)}{1-\operatorname{TPR}(A)}
\,\Big\}.
\end{equation}

In particular, any valid pure-DP guarantee must satisfy
$\varepsilon \geq \varepsilon_{\mathrm{witness}}(A).$

\textbf{Auditing setup.}
We assume that the auditor knows that the original graph is one of two edge-neighbouring graphs $G\sim G'$, differing only in a target edge $e^\star=(u,v)$. We fix $G$ (containing $e^\star$) as the positive input and $G'$ as the negative input.  The auditor has full knowledge of and query access to $\MM$. As in \cite{houssiau2022a}, the auditor repeatedly and independently applies $\mathcal{M}$ to $G$ and $G'$ to obtain $\widetilde{G}_1,\ldots,\widetilde{G}_m
\overset{\mathrm{i.i.d.}}{\sim}P_{\epsilon}(\cdot\mid G)$, and $\widetilde{G}'_1,\ldots,\widetilde{G}'_m
\overset{\mathrm{i.i.d.}}{\sim}P_{\epsilon}(\cdot\mid G')$, yielding the balanced auditing dataset
$\mathcal{D} = \{(\widetilde{G}_i,1)\}_{i=1}^{m}
\cup
\{(\widetilde{G}'_i,0)\}_{i=1}^{m}.$ 
Following empirical DP auditing
\citep{jagielski2020auditing,nasr2021adversary,nasr2023tight}, for a privatised graph $H$, the auditor learns a scoring function $s(H)$ to distinguish the two induced graph distributions. For a threshold $\tau$, let $A_\tau=\{H:s(H)\geq\tau\}$, with $\operatorname{TPR}(\tau)=P_{\epsilon}(A_\tau\mid G)$, and $\operatorname{FPR}(\tau)=P_{\epsilon}(A_\tau\mid G')$. The \textbf{empirical privacy loss witnessed for $A_\tau$ at a fixed threshold $\tau$} is given by $\epsilon_{\mathrm{audit}}(\tau)$, which estimates $\epsilon_{\mathrm{witness}}(\tau)$ by replacing TPR and FPR by their empirical estimates $\widehat{\operatorname{TPR}}$ and $\widehat{\operatorname{FPR}}$ in \eqref{eq:epswitt}; 
  \[
 {\epsilon}_{\mathrm{audit}}(\tau)
  = \max\Bigg\{\, 0,\,
\log\frac{\widehat{\operatorname{TPR}}(\tau)}
  {\widehat{\operatorname{FPR}}(\tau)},\, 
  \log\frac{1-\widehat{\operatorname{FPR}}(\tau)}
  {1-\widehat{\operatorname{TPR}}(\tau)},\,
  \Bigg\}.
  \]

Since $\operatorname{TPR}$ and $\operatorname{FPR}$ are estimated from a finite audit set, plug-in estimates can be unstable, particularly near zero. Following DP auditing practice \cite{nasr2021adversary,nasr2023tight}, we use exact one-sided Clopper--Pearson bounds $L_{\mathrm{TPR}}$ and $U_{\mathrm{FPR}}$, and report the \textbf{statistically valid lower bound}
\begin{equation}
\epsilon_{\mathrm{LB}}
=
\max\left\{\, 
0,\,
\log\frac{L_{\mathrm{TPR}}}{U_{\mathrm{FPR}}}, \, 
\log\frac{1-U_{\mathrm{FPR}}}{1-L_{\mathrm{TPR}}}\, 
\label{eq:eps_lb}
\right\}.
\end{equation}
We Bonferroni-split the error probability to give the desired simultaneous confidence level. 
Details can be found in Appendix \ref{sec:stat_bounds}. As is usual in empirical DP auditing (see, for example, \cite{nasr2023tight}), both ${\epsilon}_{\mathrm{audit}}(\tau)$ and $\epsilon_{\mathrm{LB}}$ 
measure privacy loss \emph{witnessed by the auditor}, not the true privacy parameter of the mechanism. A large witnessed loss provides evidence of leakage, whereas failure to find one does not certify the nominal (claimed) DP guarantee.

\section{Attack Models}

We construct three increasingly expressive attacks that, given an edge $e^\star$, distinguish between the distribution of privatised graphs induced by mechanism $\mathcal{M}$ using the $e^\star$-present graph $G$ and the $e^\star$-absent graph $G'$. All attacks are trained/tuned on training/validation data and evaluated on a common audit set using $\epsilon_{\mathrm{LB}}$ (eq.~\ref{eq:eps_lb}).

\textbf{Target-edge attack.}
The simplest attack uses only the observed state of the target edge $e^*$ and outputs $1$ exactly when $e^* \in E(H)$. This attack provides a baseline for direct edge leakage: differences in the probability that $e^\star$ appears under each graph distribution can directly reveal its original membership.

\textbf{Local geometric attack.}
To test whether $e^\star$ leaves a footprint in its local structural geometry, we remove the target edge from $H$, defining $H^{-e^\star}=(V(H),E(H)\setminus\{e^\star\})$, and extract
\[\big[d(u),d(v),\operatorname{CN}(u,v),\operatorname{Jaccard}(u,v),
\operatorname{AA}(u,v),\operatorname{dist}(u,v),C(u),C(v)
\big]_{H^{-e^\star}},
\]
These features capture local connectivity, neighbourhood overlap, distance, and clustering;
$\operatorname{CN}$, $\operatorname{AA}$, $\operatorname{dist}$, and $C$ denote common neighbours, Adamic--Adar, shortest-path distance, and local clustering coefficient, respectively; for definitions see Appendix \ref{app:summaries}. A binary classifier (Random Forest) then distinguishes the two graph distributions from this representation. 

\textbf{GNN-based geometric attack.}
To learn local structural geometry beyond predefined statistics, we apply a GNN to $H^{-e^{\star}(k)}_{u,v}$, the union of the $k$-hop neighbourhoods of $u$ and $v$. Each node $x$ is initialised with the feature vector 
\[
[\widetilde d(x),\mathbb{I}\{x=u\},\mathbb{I}\{x=v\},
\widetilde\delta_u(x),\widetilde\delta_v(x),C(x)],
\]
comprising normalised degree, target-node indicators, normalised distances to
$u$ and $v$, and clustering coefficient computed on $H^{-e^\star(k)}_{u,v}$. A GCN learns node embeddings $h_x$, from which we construct
\[
z_{\mathrm{GNN}}
=
[(h_u+h_v), \,|h_u-h_v|, \, (h_u\odot h_v),
\,g_{\mathrm{mean}}, \,g_{\mathrm{max}}, \,\operatorname{CN}(u,v)],
\]
where $g_{\mathrm{mean}}$ \& $g_{\mathrm{max}}$ are mean/max pooled subgraph embeddings. An MLP maps $z_{\mathrm{GNN}}$ to the attack score, and the model is trained end-to-end using binary cross-entropy.

\begin{figure}[t]
    \centering
    \includegraphics[width=\linewidth]{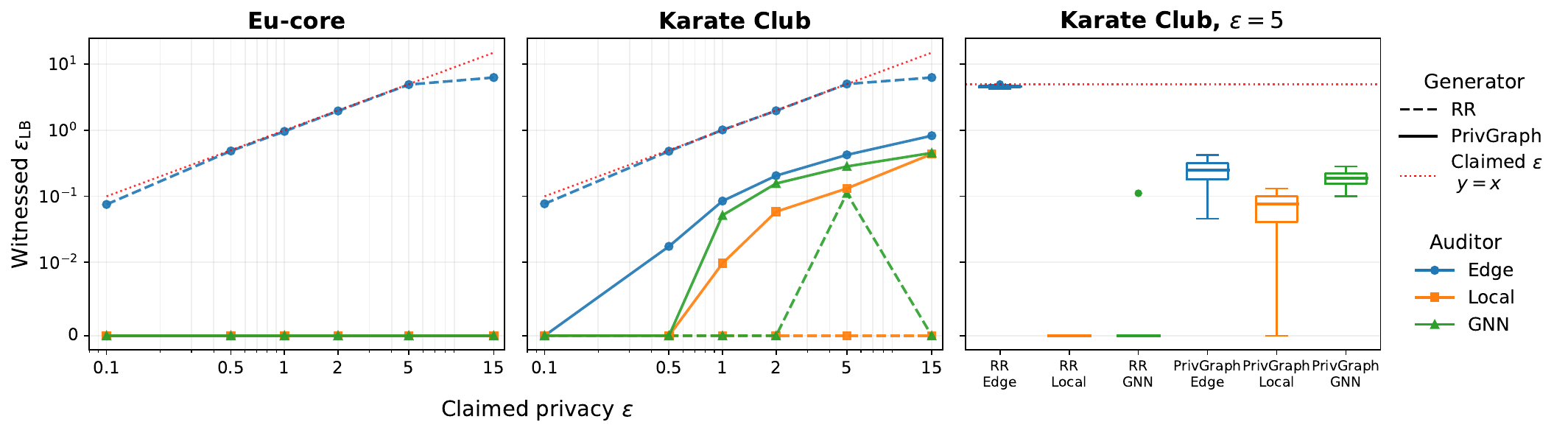}
    \caption{\small{Left and centre: Maximum $\epsilon_{\mathrm{LB}}$ across 30 randomly selected target edges as a function of the nominal privacy budget. Right: $\epsilon_{LB}$ boxplot at $\epsilon=5$.
    Colours/markers denote auditors and line styles denote generators. 
    }} 
    \label{fig:epsilon_lb_comparison}
\end{figure}

\section{Experiments}
\label{sec:experiments}

We evaluate two differentially private graph generators with $\epsilon \in \{0.1,0.5,1,2,5,15\}$ namely, \textit{Randomised response (RR)} \citep{karwa2014}, which independently perturbs potential edges, and \textit{PrivGraph} \citep{Quan2023}, which uses a private community-based graph reconstruction, on two datasets: \textit{Zachary's Karate Club}, a social network with 34 nodes and 78 edges, from \cite{zachary_information_1977}, and  \textit{Eu-Core} \citep{leskovec2007graph}, an email network from a European institution from which we take the largest department, symmetrise and retain the largest connected component ($101$ nodes, $745$ edges). Further details on the data sets and generators are in Appendix \ref{sec:extended_Resuts} and \ref{sec:generators}.

\textbf{Implementation.}
For each setting, we sample $N=30$ target edges and generate $m=10{,}000$ releases per neighbouring graph distribution. We use a stratified $60/20/20\%$ training/validation/audit split, shared across all attacks: classifiers are trained on the training split; hyperparameters, score orientation, and threshold $\tau$ are selected using only validation data; the resulting test is then frozen and evaluated once on the audit split. We report ROC--AUC and $\epsilon_{\mathrm{LB}}$, and measure structural utility by average normalised Weisfeiler--Lehman (WL) similarity across the neighbouring graphs; see Appendix \ref{app:wl_utility} for details. Our PrivGraph implementation is adapted from the implementation released with PGB \citep{liu2024pgb}; the underlying PrivGraph method is due to \citet{Quan2023}.

\paragraph{Witnessed Privacy Loss Across Networks and Privacy Levels}
We first examine how witnessed privacy loss varies with the nominal (claimed) privacy budget. Fig.~\ref{fig:epsilon_lb_comparison} reports maximum $\epsilon_{\mathrm{LB}}$ across 30 target edges sampled without replacement, with the diagonal indicating the nominal privacy level. The maximum $\epsilon_{\mathrm{LB}}$ across the 30 audited edges is a worst-case empirical summary; individual $\epsilon_{\mathrm{LB}}$ values are confidence-valid lower bounds. We find that for RR, the Edge auditor closely tracks the nominal $\epsilon$ on both networks, while the structural auditors witness little leakage. In contrast, PrivGraph exhibits smaller but increasing structural leakage on Karate Club, whereas no leakage is witnessed on the sampled EU-core edges. At large $\epsilon$, RR saturates at $\epsilon_{\mathrm{LB}} \approx 6.3$; see Appendix \ref{kclub_main} for details.

\paragraph{What Information Drives Privacy Leakage?}
We compare Edge, Local, and GNN auditors to test whether the distinguishing signal lies in the target edge itself or persists in its local structural geometry, and whether learned representations capture evidence missed by handcrafted descriptors. The result, given in Table~\ref{tab:privacy-audit}, shows a clear distinction between direct and structural leakage. For RR, the Edge auditor closely tracks the nominal $\epsilon$ on both networks, whereas Local and GNN yield $\epsilon_{\mathrm{LB}}\approx0$, indicating little detectable leakage once the target edge itself is removed. For PrivGraph, leakage is network dependent: all auditors yield $\epsilon_{\mathrm{LB}}\approx0$ on EU-core, while on Karate Club the Edge, Local, and GNN witnesses increase with $\epsilon$, suggesting PrivGraph's reconstruction can retain information about the target edge in the surrounding neighbourhood on some networks. However, since only $30$ of the $745$ EU-core edges are audited, these results do not rule out stronger leakage among unsampled edges. WL similarity generally increases with \(\epsilon\), most clearly for RR, while PrivGraph shows a flatter, network-dependent utility profile, reflecting the structural changes introduced by reconstruction.

\newlength{\withinmeth}
\setlength{\withinmeth}{1mm}
\newlength{\betmeth}
\setlength{\betmeth}{2mm}

\begin{table}[t]
\centering
\footnotesize
\setlength{\tabcolsep}{2.5pt}
\renewcommand{\arraystretch}{0.90}
\caption{Privacy utility (WL; mean$\pm$std) and audit ($\epsilon_{LB}$; mean(max)) over 30 sampled edges.}
\label{tab:privacy-audit}

\begin{tabular}{lllcccccc}
\toprule
Network & Generator & Metric
& $\epsilon$ = .1 & .5 & 1 & 2 & 5 & 15 \\
\midrule

& \multirow{4}{*}{RR}
& WL
& 0.700$\pm$0.000 & 0.701$\pm$0.000 & 0.701$\pm$0.000 & 0.709$\pm$0.000 & 0.782$\pm$0.002 & 1.000$\pm$0.000 \\
&
& Edge
& 0.022(0.077) & 0.412(0.482) & 0.925(1.014) & 1.888(1.975)	& 4.585(5.024) & 6.295(6.295) \\
&
& Local
& 0.000(0.000) & 0.000(0.000) & 0.000(0.000) & 0.000(0.000) & 0.000(0.000) & 0.000(0.000)\\
Karate & & GNN
& 0.000(0.000) & 0.000(0.000) & 0.000(0.000) & 0.000(0.000) & 0.004(0.111)
 & 0.000(0.000)\\

\cmidrule(lr){2-9}

Club &
\multirow{4}{*}{PrivGraph}
& WL
& 0.503$\pm$0.000 & 0.643$\pm$0.000 & 0.686$\pm$0.001 & 0.702$\pm$0.000 & 0.708$\pm$0.000 & 0.708$\pm$0.000 \\
&
& Edge
& 0.000(0.000) & 0.001(0.017)	& 0.003(0.085) & 0.068(0.205)	& 0.251(0.425) & 0.407(0.828) \\
&
& Local
& 0.000(0.000) & 0.000(0.000)	& 3.3$\times e^{-04}$(0.01) & 0.006(0.059)	& 0.071(0.132) & 0.198(0.437) \\
&
& GNN
& 0.000(0.000) & 0.000(0.000)	& 0.008(0.052) & 0.050(0.156)	& 0.192(0.284) & 0.317(0.455) \\

\midrule

& \multirow{4}{*}{RR}
& WL
& 0.916$\pm$0.000 & 0.922$\pm$0.000 & 0.929$\pm$0.000 & 0.943$\pm$0.000 & 0.958$\pm$0.000 & 1.000$\pm$0.000 \\
&
& Edge
& 0.025(0.075) & 0.416(0.487) & 0.914(0.963) & 1.895(1.964) & 4.586(4.923) & 6.295(6.295) \\
&
& Local
& 0.000(0.000) & 0.000(0.000) & 0.000(0.000) & 0.000(0.000) & 0.000(0.000) & 0.000(0.000) \\
EU &
& GNN
& 0.000(0.000) & 0.000(0.000) & 0.000(0.000) & 0.000(0.000) & 0.000(0.000) & 0.000(0.000) \\

\cmidrule(lr){2-9}

Core &
\multirow{4}{*}{PrivGraph}
& WL
& 0.926$\pm$0.001 & 0.937$\pm$0.000 & 0.939$\pm$0.000 & 0.941$\pm$0.000 & 0.946$\pm$0.000 & 0.951$\pm$0.000 \\
&
& Edge
& 0.000(0.000) & 0.000(0.000) & 0.000(0.000) & 0.000(0.000) & 0.000(0.000) & 0.000(0.000) \\
&
& Local
& 0.000(0.000) & 0.000(0.000) & 0.000(0.000) & 0.000(0.000) & 0.000(0.000) & 0.000(0.000) \\
&
& GNN
& 0.000(0.000) & 0.000(0.000) & 0.000(0.000) & 0.000(0.000) & 0.000(0.000) & 0.000(0.000) \\

\bottomrule
\end{tabular}
\end{table}

Summarising, RR can be used as a negative control: after removing the target edge, the remaining graph should not retain information about the target membership, as supported by the results in Tab.~\ref{tab:privacy-audit}. PrivGraph serves as the structural case; we find that reconstruction from noisy structural summaries can leave information about the target edge encoded in its local surroundings.
Moreover, handcrafted summaries may miss part of this signal, while a learned representation such as a GNN can recover it. 

\section{Discussion}

This paper finds that different mechanisms distribute the information about a neighbouring edge differently, and learned structural representations can recover leakage that fixed local summaries fail to expose. This framework for auditing DP graph generation can help developers detect implementation
errors, quantify attack-accessible privacy loss, and communicate the limits of private graph releases more
accurately (with a risk of abuse by attackers). There are some possible extensions. While we use  ``pure'' $\epsilon$-edge DP, other notions of differential privacy are available; see, for example, \cite{wang2020comprehensive}. The reported $\epsilon_{\mathrm{LB}}$ is attack-dependent and represents only the privacy loss witnessed by the auditor; failure to exceed the nominal $\epsilon$ does not certify the DP guarantee of the mechanism. Our experiments consider single-release inference, whereas multiple independent releases may enable stronger attacks through evidence aggregation. Finally, evaluation on a small set of networks and attacks does not establish generality across graph geometries or adversarial strategies; broader graph families, stronger attacks see, for example, \cite{xiangfei2025defending}, and repeated-release auditing are left for future work.

\begin{ack}
AF and AE are supported by UKRI EPSRC grant EP/V056883/1.
GR is supported in part by UKRI EPSRC grants EP/T018445/1,   EP/V056883/1, EP/X002195/1 and  EP/Y028872/1. 
LS is supported in part by UKRI EPSRC grant EP/V056883/1.
CM is supported in part by UKRI EPSRC grants EP/V056883/1 and EP/R007195/1. 
\end{ack}

\bibliography{references.bib}
\bibliographystyle{plainnat}


\newpage

\appendix
\startcontents[appendices]
\printcontents[appendices]{}{1}{\section*{Appendix: Table of Contents}}
\clearpage 

\section{Notation}

Table \ref{tab:notation} lists the main notation used in this paper.  

\begin{table}[h]
\centering
\caption{Summary of notation used throughout the paper.}
\label{tab:notation}
\begin{tabular}{ll}\\
\toprule
\textbf{Notation} & \textbf{Description} \\
\midrule

$G=(V,E)$
& Original input graph \\

$G'$
& Edge-neighbouring graph of $G$ \\

$G\sim G'$
& $G$ and $G'$ differ in exactly one edge \\

$e^\star=(u,v)$
& Target edge distinguishing $G$ and $G'$ \\

$\mathcal{M}$
& Randomised private graph-generation mechanism \\

$\epsilon$
& Differential privacy parameter \\

$P_{\epsilon}(\cdot\mid G)$
& Distribution induced by $\mathcal{M}$ on input $G$ \\

$H$
& Privatised graph released by $\mathcal{M}$ \\

$H^{-e^\star}$
& $H$ with the target edge $e^\star$ removed \\

$Y$
& edge presence label; $Y=1$ for $G$ and $Y=0$ for $G'$ \\
$s(H)$
& Auditor score for a privatised graph $H$ \\

$\tau$
& Decision threshold for the auditor \\

$A_\tau$
& Positive decision region $\{H:s(H)\geq\tau\}$ \\

$\operatorname{TPR},\operatorname{FPR}$
& True- and false-positive rates of the auditor \\

$\epsilon_{witness}$ & Privacy loss witnessed by the auditor for a decision event \\

$\epsilon_{\mathrm{audit}}$
& Empirical privacy loss witnessed by the auditor \\

$\epsilon_{\mathrm{LB}}$
& Confidence-valid lower bound on witnessed privacy loss \\

$L_{\mathrm{TPR}},U_{\mathrm{TPR}}$
& Lower/upper confidence bounds on TPR \\

$L_{\mathrm{FPR}},U_{\mathrm{FPR}}$
& Lower/upper confidence bounds on FPR \\

$H^{-e^\star(k)}_{u,v}$
& Edge-removed target-centred $k$-hop subgraph \\

$\widetilde d(x)$
& Degree of $x$ normalised by maximum subgraph degree \\

$\widetilde\delta_u(x),\widetilde\delta_v(x)$
& Distances from $x$ to $u,v$, scaled by $k$ \\

$C(x)$
& Clustering coefficient of $x$ in the local subgraph \\

$\operatorname{CN}(u,v)$
& Number of common neighbours of $u$ and $v$ \\

$h_x$
& GNN-learned representation of node $x$ \\

$g_{\mathrm{mean}},g_{\mathrm{max}}$
& Mean/max pooled local-subgraph representations \\

$z_{\mathrm{GNN}}$
& Final representation used by the GNN auditor \\

$f_\omega$
& Binary classifier applied to the GNN representation \\

$m$
& Number of privatised samples generated per neighbouring graph
\\

$N$
& Number of sampled target edges \\

\bottomrule
\end{tabular}
\end{table}

\clearpage 
\section{Additional Details and Results for the Experiments} \label{sec:extended_Resuts}

In this section, we provide details on the real-world networks used in the main paper's experiments and present supplementary results for those experiments. Alongside the ones reported in the main paper, we report the following additional audit metrics

\paragraph{Additional audit metrics.}
We report the so-called {\it membership-inference advantage} 
\[\mathrm{MIA \, advantage} =
\widehat{\operatorname{TPR}}
-
\widehat{\operatorname{FPR}},
\]
which measures the auditor's distinguishing advantage at the selected threshold. We also report the audit tightness
\[
\operatorname{Tightness}
=
\frac{\epsilon_{\mathrm{LB}}}{\epsilon},
\]
which measures the fraction of the nominal privacy budget witnessed by the audit.

\subsection{Zachary's Karate Club Network} \label{kclub_main}
Zachary’s karate club network (\cite{zachary_information_1977})  is a friendship network of 34 members of a karate club at a US university, which split into two factions as a result of an internal dispute. This network is often used as a benchmark network dataset for community detection algorithms; see, for example, \cite{Barbour_Reinert_2026}.

We use the Karate Club network as a small, well-characterised benchmark for auditing privacy mechanisms and for assessing structural utility. We randomly sample $30$ existing edges as audit targets. For each target edge $e^\star=(u,v)$, we construct a pair of edge-neighbouring graphs by retaining $e^\star$ in one graph and removing it in the other. For each privacy level $\epsilon \in\{0.1,0.5,1,2,5,15\}$, we apply each graph privatisation mechanism independently to the two neighbouring graphs and evaluate the resulting samples using the Edge, Local, and GNN auditors described in the main text. Table~\ref{tab:privacy-results} reports the corresponding detailed privacy-auditing and structural-utility results along with the additional measures computed for the audit.

The RR mechanism tends to give higher utility than PrivGraph, when measured by the WL similarity. The witness bound $\epsilon_{LB}$ is almost always considerably lower than the nominal $\epsilon$, except for the Edge attack in RR, for which it is very close to the nominal value (as expected, due to the RR mechanism). An exception occurs for $\epsilon=15$; for this small network, under RR,  6.295 is the largest theoretically possible value: if the classifier is perfect, with TPR equal 1 and FPR equal 0, for $m=2000$ samples and $\alpha = 0.05$, the Clopper-Pearson bound would be 
$ \log \left( \frac{(0.025)^{1/2000}}{ 1 - (0.025)^{1/2000}}\right) \approx 6.295.$ The MIA advantage of 1 reflects this situation.

For PrivGraph, even the edge attack gives a witness that is well below the nominal $\epsilon$, even taking random variation into account. For fixed $\epsilon$, under PrivGraph the GNN attacks tend to yield the highest MIA advantage.

\begin{figure}[t]
    \centering
    \includegraphics[width=\linewidth]{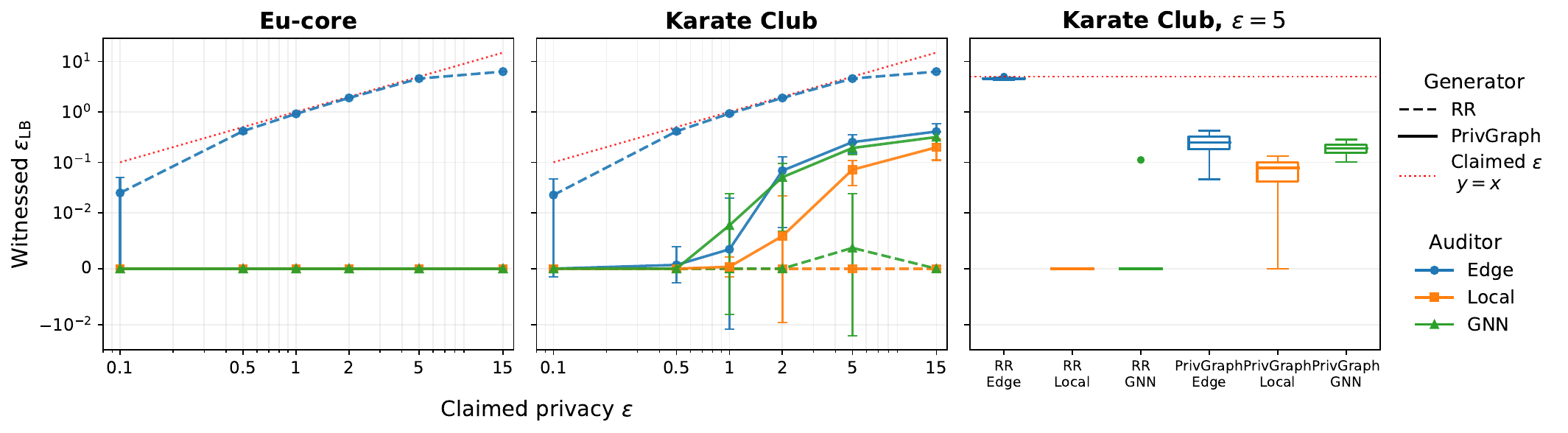}
    \caption{\small{Left and centre: Average $\epsilon_{\mathrm{LB}}$ across 30 randomly selected target edges as a function of the nominal privacy budget. Right: $\epsilon_{LB}$ boxplot at $\epsilon=5$. Colours/markers denote auditors and line styles denote generators.} } 
    \label{fig:epsilon_lb_comparison_avrg}
\end{figure}

\begingroup
\scriptsize
\setlength{\tabcolsep}{2.5pt}
\renewcommand{\arraystretch}{0.90}

\begin{longtable}{l l l c c c c c c c c c}

\caption{Privacy auditing and structural utility results across graph
generation mechanisms and attack models for the Karate Club Network.}
\label{tab:privacy-results}\\

\toprule
Gen. & Aud. & $\epsilon$
& WL Utility
& \multicolumn{2}{c}{$\epsilon_{\mathrm{LB}}$}
& \multicolumn{2}{c}{Audit Tightness}
& \multicolumn{2}{c}{$\epsilon_{\mathrm{audit}}$}
& AUC
& MIA Advantage
\\
\multicolumn{3}{c}{}
& (Mean $\pm$ Std)
& (Mean $\pm$ Std)
& (Max)
& (Mean $\pm$ Std)
& (Max)
& (Mean $\pm$ Std)
& (Max)
& (Mean $\pm$ Std)
& (Mean $\pm$ Std)
\\
\midrule
\endfirsthead

\toprule
Gen. & Aud. & $\epsilon$
& WL Utility
& \multicolumn{2}{c}{$\epsilon_{\mathrm{LB}}$}
& Audit Tightness
& \multicolumn{2}{c}{$\epsilon_{\mathrm{audit}}$}
& AUC
& MIA Advantage
\\
\multicolumn{3}{c}{}
& (Mean $\pm$ Std)
& (Mean $\pm$ Std)
& (Max)
& (Mean $\pm$ Std)
& (Max)
& (Mean $\pm$ Std)
& (Max)
& (Mean $\pm$ Std)
& (Mean $\pm$ Std)
\\
\midrule
\endhead

\midrule
\multicolumn{11}{r}{Continued on next page}\\
\endfoot

\bottomrule
\endlastfoot

\multirow{20}{*}{\textbf{RR}}
& \multirow{6}{*}{Edge}
& 0.1 & 0.699$\pm$0.002 & 0.022$\pm$0.024 & 0.077 & 0.224$\pm$0.239 & 0.771 & 0.106$\pm$0.034 & 0.167 & 0.526$\pm$0.008 & 0.052$\pm$0.017 \\
&& 0.5 & 0.699$\pm$0.002 & 0.412$\pm$0.033 & 0.482 & 0.824$\pm$0.066 & 0.965 & 0.505$\pm$0.033 & 0.576 & 0.622$\pm$0.008 & 0.243$\pm$0.016 \\
&& 1 & 0.700$\pm$0.002 & 0.925$\pm$0.046 & 1.014 & 0.925$\pm$0.046 & 1.014 & 1.027$\pm$0.047 & 1.118 & 0.733$\pm$0.009 & 0.466$\pm$0.017 \\
&& 2 & 0.708$\pm$0.002 & 1.888$\pm$0.053 & 1.975 & 0.944$\pm$0.027 & 0.988 & 2.027$\pm$0.057 & 2.119 & 0.880$\pm$0.006 & 0.760$\pm$0.011 \\
&& 5 & 0.779$\pm$0.002 & 4.585$\pm$0.174 & 5.024 & 0.917$\pm$0.035 & 1.005 & 5.164$\pm$0.242 & 5.804 & 0.993$\pm$0.001 & 0.987$\pm$0.003 \\
&& 15 & 1.000$\pm$0.000 & 6.295$\pm$0.000 & 6.295 & 0.420$\pm$0.000 & 0.42 & — & inf & 1.000$\pm$0.000 & 1.000$\pm$0.000 \\
\\
&\multirow{6}{*}{Local}
& 0.1 & 0.699$\pm$0.002 & 0.000$\pm$0.000 & 0.000 & 0.000$\pm$0.000 & 0.000 & 0.036$\pm$0.059 & 0.251 & 0.501$\pm$0.010 & 0.001$\pm$0.012 \\
&& 0.5 & 0.699$\pm$0.002 & 0.000$\pm$0.000 & 0.000 & 0.000$\pm$0.000 & 0.000 & 0.030$\pm$0.039 & 0.135 & 0.501$\pm$0.009 & 0.001$\pm$0.014 \\
&& 1 & 0.700$\pm$0.002 & 0.000$\pm$0.000 & 0.000 & 0.000$\pm$0.000 & 0.000 & 0.033$\pm$0.057 & 0.288 & 0.500$\pm$0.008 & 0.002$\pm$0.013 \\
&& 2 & 0.708$\pm$0.002 & 0.000$\pm$0.000 & 0.000 & 0.000$\pm$0.000 & 0.000 & 0.041$\pm$0.088 & 0.477 & 0.501$\pm$0.011 & 0.001$\pm$0.015 \\
&& 5 & 0.779$\pm$0.002 & 0.000$\pm$0.000 & 0.000 & 0.000$\pm$0.000 & 0.000 & 0.068$\pm$0.118 & 0.47 & 0.504$\pm$0.010 & 0.004$\pm$0.012 \\
&& 15 & 1.000$\pm$0.000 & 0.000$\pm$0.000 & 0.000 & 0.000$\pm$0.000 & 0.000 & 0.000$\pm$0.000 & 0.000 & 0.500$\pm$0.000 & -0.000$\pm$0.000 \\
\\
& \multirow{5}{*}{GNN}
& 0.1 & 0.699$\pm$0.002 & 0.000$\pm$0.000 & 0.000 & 0.000$\pm$0.000 & 0.000 & 0.051$\pm$0.127 & 0.693 & 0.502$\pm$0.010 & 0.004$\pm$0.014 \\
&& 0.5 & 0.699$\pm$0.002 & 0.000$\pm$0.000 & 0.000 & 0.000$\pm$0.000 & 0.000 & 0.032$\pm$0.059 & 0.215 & 0.498$\pm$0.010 & -0.001$\pm$0.015 \\
&& 1 & 0.700$\pm$0.002 & 0.000$\pm$0.000 & 0.000 & 0.000$\pm$0.000 & 0.000 & — & inf & 0.497$\pm$0.011 & -0.002$\pm$0.014 \\
&& 2 & 0.708$\pm$0.002 & 0.000$\pm$0.000 & 0.000 & 0.000$\pm$0.000 & 0.000 & — & inf & 0.499$\pm$0.008 & -0.002$\pm$0.014 \\
&& 5 & 0.779$\pm$0.002 & 0.004$\pm$0.020 & 0.111 & 0.001$\pm$0.004 & 0.022 & — & inf & 0.500$\pm$0.010 & 0.001$\pm$0.007 \\
&& 15 & 1.000$\pm$0.000 & 0.000$\pm$0.000 & 0.000 & 0.000$\pm$0.000 & 0.000 & 0.000$\pm$0.000 & 0.000 & 0.500$\pm$0.001 & 0.000$\pm$0.000 \\

\midrule


& \multirow{6}{*}{Edge}
& 0.1 & 0.500$\pm$0.002 & 0.000$\pm$0.000 & 0.000 & 0.000$\pm$0.000 & 0.000 & 0.034$\pm$0.055 & 0.229 & 0.500$\pm$0.003 & -0.001$\pm$0.006 \\
&& 0.5 & 0.641$\pm$0.002 & 0.001$\pm$0.003 & 0.017 & 0.001$\pm$0.006 & 0.035 & 0.088$\pm$0.092 & 0.299 & 0.504$\pm$0.006 & 0.008$\pm$0.012 \\
&& 1 & 0.684$\pm$0.002 & 0.003$\pm$0.016 & 0.085 & 0.003$\pm$0.016 & 0.085 & 0.160$\pm$0.081 & 0.373 & 0.511$\pm$0.005 & 0.021$\pm$0.010 \\
&& 2 & 0.701$\pm$0.002 & 0.068$\pm$0.061 & 0.205 & 0.034$\pm$0.030 & 0.103 & 0.259$\pm$0.075 & 0.456 & 0.522$\pm$0.010 & 0.045$\pm$0.019 \\
&& 5 & 0.706$\pm$0.002 & 0.251$\pm$0.097 & 0.425 & 0.050$\pm$0.019 & 0.085 & 0.441$\pm$0.091 & 0.646 & 0.545$\pm$0.019 & 0.091$\pm$0.038 \\
&& 15 & 0.706$\pm$0.002 & 0.407$\pm$0.173 & 0.828 & 0.027$\pm$0.012 & 0.055 & 0.604$\pm$0.205 & 1.079 & 0.557$\pm$0.020 & 0.114$\pm$0.039 \\
\\
&\multirow{6}{*}{Local}
& 0.1 & 0.500$\pm$0.002 & 0.000$\pm$0.000 & 0.000 & 0.000$\pm$0.000 & 0.000 & 0.021$\pm$0.042 & 0.188 & 0.498$\pm$0.009 & -0.004$\pm$0.011 \\
&& 0.5 & 0.641$\pm$0.002 & 0.000$\pm$0.000 & 0.000 & 0.000$\pm$0.000 & 0.000 & 0.045$\pm$0.066 & 0.316 & 0.505$\pm$0.008 & 0.007$\pm$0.012 \\
\textbf{Priv}&& 1 & 0.684$\pm$0.002 & 0.000$\pm$0.002 & 0.01 & 0.000$\pm$0.002 & 0.01 & 0.052$\pm$0.079 & 0.405 & 0.506$\pm$0.010 & 0.008$\pm$0.016 \\
\textbf{Graph}&& 2 & 0.701$\pm$0.002 & 0.006$\pm$0.015 & 0.059 & 0.003$\pm$0.008 & 0.029 & 0.079$\pm$0.054 & 0.215 & 0.520$\pm$0.012 & 0.027$\pm$0.019 \\
&& 5 & 0.706$\pm$0.002 & 0.071$\pm$0.037 & 0.132 & 0.014$\pm$0.007 & 0.026 & 0.180$\pm$0.044 & 0.259 & 0.549$\pm$0.008 & 0.072$\pm$0.015 \\
&& 15 & 0.706$\pm$0.002 & 0.198$\pm$0.088 & 0.437 & 0.013$\pm$0.006 & 0.029 & 0.300$\pm$0.086 & 0.564 & 0.587$\pm$0.022 & 0.131$\pm$0.039 \\
\\
& \multirow{6}{*}{GNN}
& 0.1 & 0.500$\pm$0.002 & 0.000$\pm$0.000 & 0.000 & 0.000$\pm$0.000 & 0.000 & 0.056$\pm$0.179 & 0.956 & 0.501$\pm$0.008 & 0.001$\pm$0.012 \\
&& 0.5 & 0.641$\pm$0.002 & 0.000$\pm$0.000 & 0.000 & 0.000$\pm$0.000 & 0.000 & 0.071$\pm$0.088 & 0.452 & 0.511$\pm$0.007 & 0.013$\pm$0.013 \\
&& 1 & 0.684$\pm$0.002 & 0.008$\pm$0.016 & 0.052 & 0.008$\pm$0.016 & 0.052 & 0.099$\pm$0.057 & 0.227 & 0.522$\pm$0.011 & 0.029$\pm$0.017 \\
&& 2 & 0.701$\pm$0.002 & 0.050$\pm$0.044 & 0.156 & 0.025$\pm$0.022 & 0.078
& 0.159$\pm$0.060 & 0.28 & 0.542$\pm$0.010 & 0.059$\pm$0.018 \\
&& 5 & 0.706$\pm$0.002 & 0.192$\pm$0.051 & 0.284 & 0.038$\pm$0.010 & 0.057 & 0.300$\pm$0.059 & 0.408 & 0.582$\pm$0.011 & 0.120$\pm$0.018 \\
&& 15 & 0.706$\pm$0.002 & 0.317$\pm$0.081 & 0.455 & 0.021$\pm$0.005 & 0.03 & 0.419$\pm$0.083 & 0.568 & 0.625$\pm$0.023 & 0.181$\pm$0.034

\end{longtable}

\endgroup

\subsection{EU-core Network}
\label{eu_core_main}

The Email-Eu-core network \citep{leskovec2007graph} is derived from email communication at a large European research institution. The original data contain anonymised records of incoming and outgoing emails over 18 months, from October 2003 to May 2005, recording the time, sender, and recipient of each email. Following the construction of \citet{kossinets_empirical_2006}, an undirected edge is placed between two email addresses only when communication is reciprocal, i.e., each address has sent at least one email to the other.

The institutional email addresses are associated with organisational departments, providing a natural community structure. For our experiments, we restrict the network to the largest department and retain its largest connected component. This yields an undirected graph with $101$ nodes and $745$ edges. We refer to this graph as Email-Eu-core throughout the experiments.

As for the Karate Club experiment, we randomly sample 30 existing edges as audit targets. For each target edge $e^\star=(u,v)$, we construct a pair of edge-neighbouring graphs by retaining $e^\star$ in one graph and removing it in the other. For each nominal privacy level $\epsilon \in\{0.1,0.5,1,2,5,15\}$, we independently apply each graph privatisation mechanism to the two neighbouring graphs and evaluate the resulting samples using the Edge, Local, and GNN auditors described in the main text. Table \ref{tab:privacy-results_EU-core} reports the corresponding privacy-auditing and structural-utility results.

Again, for RR, the Edge attack tends to recover the nominal $\epsilon$ value well, while the other attacks give a witness lower bound which is often 0. Indeed, the Clopper-Pearson endpoints in $\epsilon_{LB}$ give log ratios below zero because the lower confidence bound on the TPR does not exceed the upper confidence bound on the FPR.

\begingroup
\scriptsize
\setlength{\tabcolsep}{2.5pt}
\renewcommand{\arraystretch}{0.90}

\begin{longtable}{l l l c c c c c c c c c}

\caption{Privacy auditing and structural utility results across graph generation mechanisms and attack models for the EU-core Network.}
\label{tab:privacy-results_EU-core}\\

\toprule
Gen. & Aud. & $\epsilon$
& WL Utility
& \multicolumn{2}{c}{$\epsilon_{\mathrm{LB}}$}
& \multicolumn{2}{c}{Audit Tightness}
& \multicolumn{2}{c}{$\epsilon_{\mathrm{audit}}$}
& AUC
& MIA Advantage
\\
\multicolumn{3}{c}{}
& (Mean $\pm$ Std)
& (Mean $\pm$ Std)
& (Max)
& (Mean $\pm$ Std)
& (Max)
& (Mean $\pm$ Std)
& (Max)
& (Mean $\pm$ Std)
& (Mean $\pm$ Std)
\\
\midrule
\endfirsthead
\toprule
Gen. & Aud. & $\epsilon$
& WL Utility
& \multicolumn{2}{c}{$\epsilon_{\mathrm{LB}}$}
& Audit Tightness
& \multicolumn{2}{c}{$\epsilon_{\mathrm{audit}}$}
& AUC
& MIA Advantage
\\
\multicolumn{3}{c}{}
& (Mean $\pm$ Std)
& (Mean $\pm$ Std)
& (Max)
& (Mean $\pm$ Std)
& (Max)
& (Mean $\pm$ Std)
& (Max)
& (Mean $\pm$ Std)
& (Mean $\pm$ Std)
\\
\midrule
\endhead
\midrule
\multicolumn{11}{r}{Continued on next page}\\
\endfoot
\bottomrule
\endlastfoot

\multirow{20}{*}{\textbf{RR}}
& \multirow{6}{*}{Edge}
& 0.1 & 0.916$\pm$0.000 & 0.025$\pm$0.025 & 0.075 & 0.247$\pm$0.252 & 0.754 & 0.107$\pm$0.035 & 0.165 & 0.526$\pm$0.009 & 0.053$\pm$0.017 \\
&& 0.5 & 0.922$\pm$0.000 & 0.416$\pm$0.035 & 0.487 & 0.833$\pm$0.069 & 0.973 & 0.509$\pm$0.035 & 0.58 & 0.623$\pm$0.008 & 0.246$\pm$0.016 \\
&& 1 & 0.929$\pm$0.000 & 0.914$\pm$0.030 & 0.963 & 0.914$\pm$0.030 & 0.963 & 1.015$\pm$0.031 & 1.064 & 0.731$\pm$0.006 & 0.462$\pm$0.012 \\
&& 2 & 0.943$\pm$0.000 & 1.895$\pm$0.035 & 1.964 & 0.948$\pm$0.017 & 0.982 & 2.034$\pm$0.037 & 2.106 & 0.881$\pm$0.004 & 0.762$\pm$0.008 \\
&& 5 & 0.958$\pm$0.000 & 4.586$\pm$0.147 & 4.923 & 0.917$\pm$0.029 & 0.985 & 5.165$\pm$0.205 & 5.649 & 0.993$\pm$0.001 & 0.987$\pm$0.002 \\
&& 15 & 1.000$\pm$0.000 & 6.295$\pm$0.000 & 6.295 & 0.420$\pm$0.000 & 0.42 & — & inf & 1.000$\pm$0.000 & 1.000$\pm$0.000 \\
\\
&\multirow{6}{*}{Local}
& 0.1 & 0.916$\pm$0.000 & 0.000$\pm$0.000 & 0 & 0.000$\pm$0.000 & 0 & 0.038$\pm$0.094 & 0.495 & 0.501$\pm$0.010 & 0.001$\pm$0.014 \\
&& 0.5 & 0.922$\pm$0.000 & 0.000$\pm$0.000 & 0 & 0.000$\pm$0.000 & 0 & 0.052$\pm$0.077 & 0.357 & 0.502$\pm$0.008 & 0.005$\pm$0.011 \\
&& 1 & 0.929$\pm$0.000 & 0.000$\pm$0.000 & 0 & 0.000$\pm$0.000 & 0 & 0.030$\pm$0.043 & 0.166 & 0.502$\pm$0.011 & 0.003$\pm$0.015 \\
&& 2 & 0.943$\pm$0.000 & 0.000$\pm$0.000 & 0 & 0.000$\pm$0.000 & 0 & 0.036$\pm$0.047 & 0.154 & 0.499$\pm$0.009 & 0.001$\pm$0.012 \\
&& 5 & 0.958$\pm$0.000 & 0.000$\pm$0.000 & 0 & 0.000$\pm$0.000 & 0 & 0.024$\pm$0.040 & 0.139 & 0.500$\pm$0.008 & -0.003$\pm$0.013 \\
&& 15 & 1.000$\pm$0.000 & 0.000$\pm$0.000 & 0 & 0.000$\pm$0.000 & 0 & — & inf & 0.500$\pm$0.000 & 0.000$\pm$0.000
\\
\\
& \multirow{5}{*}{GNN}
& 0.1 & 0.916$\pm$0.000 & 0.000$\pm$0.000 & 0 & 0.000$\pm$0.000 & 0 & 0.048$\pm$0.139 & 0.693 & 0.499$\pm$0.010 & -0.002$\pm$0.011 \\
&& 0.5 & 0.922$\pm$0.000 & 0.000$\pm$0.000 & 0 & 0.000$\pm$0.000 & 0 & 0.034$\pm$0.049 & 0.149 & 0.499$\pm$0.011 & 0.002$\pm$0.014 \\
&& 1 & 0.929$\pm$0.000 & 0.000$\pm$0.000 & 0 & 0.000$\pm$0.000 & 0 & 0.023$\pm$0.057 & 0.266 & 0.497$\pm$0.008 & -0.004$\pm$0.013 \\
&& 2 & 0.943$\pm$0.000 & 0.000$\pm$0.000 & 0 & 0.000$\pm$0.000 & 0 & — & inf & 0.500$\pm$0.010 & 0.000$\pm$0.010 \\
&& 5 & 0.958$\pm$0.000 & 0.000$\pm$0.000 & 0 & 0.000$\pm$0.000 & 0 & 0.034$\pm$0.067 & 0.227 & 0.504$\pm$0.009 & 0.001$\pm$0.009 \\
&& 15 & 1.000$\pm$0.000 & 0.000$\pm$0.000 & 0 & 0.000$\pm$0.000 & 0 & 0.000$\pm$0.000 & 0 & 0.500$\pm$0.001 & 0.000$\pm$0.000 \\
\midrule


& \multirow{6}{*}{Edge}
& 0.1 & 0.926$\pm$0.000 & 0.000$\pm$0.000 & 0 & 0.000$\pm$0.000 & 0 & 0.054$\pm$0.075 & 0.241 & 0.501$\pm$0.005 & 0.002$\pm$0.011 \\
&& 0.5 & 0.937$\pm$0.000 & 0.000$\pm$0.000 & 0 & 0.000$\pm$0.000 & 0 & 0.051$\pm$0.074 & 0.227 & 0.501$\pm$0.007 & 0.001$\pm$0.014 \\
&& 1 & 0.939$\pm$0.000 & 0.000$\pm$0.000 & 0 & 0.000$\pm$0.000 & 0 & 0.022$\pm$0.034 & 0.142 & 0.499$\pm$0.004 & -0.001$\pm$0.009 \\
&& 2 & 0.941$\pm$0.000 & 0.000$\pm$0.000 & 0 & 0.000$\pm$0.000 & 0 & 0.038$\pm$0.043 & 0.132 & 0.501$\pm$0.005 & 0.002$\pm$0.011 \\
&& 5 & 0.946$\pm$0.000 & 0.000$\pm$0.000 & 0 & 0.000$\pm$0.000 & 0 & 0.034$\pm$0.057 & 0.226 & 0.498$\pm$0.006 & -0.003$\pm$0.011 \\
&& 15 & 0.951$\pm$0.000 & 0.000$\pm$0.000 & 0 & 0.000$\pm$0.000 & 0 & 0.038$\pm$0.074 & 0.265 & 0.499$\pm$0.006 & -0.001$\pm$0.011 \\
\\
&\multirow{6}{*}{Local}
& 0.1 & 0.926$\pm$0.000 & 0.000$\pm$0.000 & 0 & 0.000$\pm$0.000 & 0 & 0.028$\pm$0.056 & 0.262 & 0.499$\pm$0.010 & -0.001$\pm$0.015 \\
&& 0.5 & 0.937$\pm$0.000 & 0.000$\pm$0.000 & 0 & 0.000$\pm$0.000 & 0 & 0.064$\pm$0.135 & 0.693 & 0.503$\pm$0.010 & 0.002$\pm$0.013 \\
&& 1 & 0.939$\pm$0.000 & 0.000$\pm$0.000 & 0 & 0.000$\pm$0.000 & 0 & 0.047$\pm$0.132 & 0.693 & 0.502$\pm$0.008 & -0.002$\pm$0.011 \\
&& 2 & 0.941$\pm$0.000 & 0.000$\pm$0.000 & 0 & 0.000$\pm$0.000 & 0 & 0.028$\pm$0.035 & 0.131 & 0.499$\pm$0.010 & -0.000$\pm$0.014 \\
&& 5 & 0.946$\pm$0.000 & 0.000$\pm$0.000 & 0 & 0.000$\pm$0.000 & 0 & 0.031$\pm$0.064 & 0.336 & 0.501$\pm$0.008 & 0.002$\pm$0.013 \\
&& 15 & 0.951$\pm$0.000 & 0.000$\pm$0.000 & 0 & 0.000$\pm$0.000 & 0 & 0.025$\pm$0.087 & 0.477 & 0.500$\pm$0.008 & -0.002$\pm$0.012 \\
\\
& \multirow{6}{*}{GNN}
& 0.1 & 0.926$\pm$0.000 & 0.000$\pm$0.000 & 0 & 0.000$\pm$0.000 & 0 & 0.037$\pm$0.067 & 0.26 & 0.502$\pm$0.011 & -0.001$\pm$0.013 \\
&& 0.5 & 0.937$\pm$0.000 & 0.000$\pm$0.000 & 0 & 0.000$\pm$0.000 & 0 & 0.045$\pm$0.086 & 0.452 & 0.500$\pm$0.009 & 0.002$\pm$0.014 \\
&& 1 & 0.939$\pm$0.000 & 0.000$\pm$0.000 & 0 & 0.000$\pm$0.000 & 0 & 0.098$\pm$0.203 & 0.827 & 0.500$\pm$0.012 & 0.001$\pm$0.017 \\
&& 2 & 0.941$\pm$0.000 & 0.000$\pm$0.000 & 0 & 0.000$\pm$0.000 & 0 & 0.039$\pm$0.070 & 0.306 & 0.500$\pm$0.008 & -0.000$\pm$0.011 \\
&& 5 & 0.946$\pm$0.000 & 0.000$\pm$0.000 & 0 & 0.000$\pm$0.000 & 0 & 0.023$\pm$0.042 & 0.158 & 0.499$\pm$0.008 & 0.000$\pm$0.009 \\
&& 15 & 0.951$\pm$0.000 & 0.000$\pm$0.000 & 0 & 0.000$\pm$0.000 & 0 & 0.014$\pm$0.034 & 0.166 & 0.501$\pm$0.009 & -0.002$\pm$0.013

\end{longtable}
\endgroup

\subsection{Experimental settings and Compute Resources}
For all the experiments in our paper, we use

\begin{table}[t]
\centering
\scriptsize
\setlength{\tabcolsep}{4pt}
\renewcommand{\arraystretch}{0.92}
\caption{Implementation specification for the Local and GNN auditors and
the PrivGraph generator.}
\label{tab:implementation-spec}
\begin{tabular}{p{0.26\linewidth} p{0.68\linewidth}}
\toprule
\textbf{Component} & \textbf{Specification} \\
\midrule

\multicolumn{2}{l}{\textbf{Audit protocol}} \\
Samples per neighbouring graph & $m=10{,}000$ releases per graph; stratified
$60/20/20\%$ train/validation/audit split, giving $6000/2000/2000$ samples
per graph. \\
Shared releases & The same generated releases and split are used by all
auditors for each $(\text{generator},\epsilon,e^\star)$ setting. \\
Training and selection & Train on the training split; select model/score
orientation and threshold $\tau$ using validation data only; freeze the
resulting classifier and evaluate once on the audit split. \\
Confidence & $95\%$ simultaneous confidence; Bonferroni correction with
$\alpha/2$ allocated to each of the two one-sided TPR/FPR bounds. \\

\midrule
\multicolumn{2}{l}{\textbf{Local auditor}} \\
Classifier & Random Forest: 100 trees, maximum depth 15, parallel tree
fitting ($n_{\mathrm{jobs}}=-1$). \\
Features & Common-neighbour count, Adamic--Adar, and the other explicitly
specified local structural features in the implementation; target edge is
excluded. \\
Feature standardisation & None. \\
Infinite distances & Disconnected target endpoints are assigned $|V|+1$. \\
Regularisation & Maximum tree depth 15; no additional regularisation. \\
Solver / optimiser & Scikit-learn Random Forest; no separate optimiser. \\
Hyperparameter search & None; $n_{\mathrm{estimators}}=100$ and maximum depth $=15$ are fixed. \\
Threshold selection & Validation-only; choose $\tau$ from the unique validation predicted
probabilities to maximise $\operatorname{TPR}-\operatorname{FPR}$. \\

\midrule
\multicolumn{2}{l}{\textbf{GNN auditor}} \\
Input graph & Edge-removed target-centred $k$-hop subgraph, with $k=2$. \\
Architecture & Three GCN layers with hidden dimension 64; classifier head
$64\!\rightarrow\!32\!\rightarrow\!1$ with ReLU activations and dropout $0.2$. \\
Output & One scalar logit, converted to a probability using the sigmoid function. \\
Input features
& Six node features: normalised degree, target-$u$ indicator, target-$v$
indicator, normalised distances to $u$ and $v$, and clustering coefficient. \\
Target-edge removal
& The target edge is removed before constructing the $k$-hop subgraph;
$k=2$. \\
Unreachable distance & Distance truncated to $k+1$. \\
Pair/context features
& Target-node embeddings are combined as $[h_u+h_v,\lvert h_u-h_v\rvert,h_u\odot h_v]$, together with mean/max pooled node embeddings and the common-neighbour count. \\
Normalisation
& Degree and distance features are normalised as above; no additional
feature or batch normalisation is used. \\
Dropout & $0.2$ in the classifier head. \\
Weight decay & None specified. \\
Optimiser & Adam. \\
Learning rate & $0.001$. \\
Batch size & 64. \\
Epochs & Maximum 15. \\
Early stopping & Patience $5$ based on validation loss; the best validation-loss checkpoint
is restored. \\
Validation criterion & Binary cross-entropy with logits (\texttt{BCEWithLogitsLoss}). \\
Threshold selection
& Validation-only; choose $\tau$ from the unique validation predicted
probabilities to maximise $\operatorname{TPR}-\operatorname{FPR}$. \\
Random seed & NumPy and PyTorch are seeded per audit run using
$42+10{,}000\,\mathrm{epsilon\_id}+\mathrm{edge\_id}$, where $\mathrm{epsilon\_id}$ is the index of the nominal privacy level $\epsilon \in \{0.1,0.5,1,2,5,15\}$, and $\mathrm{edge\_id}$ is the index of the sampled target edge. \\
Software & PyTorch: 1.12.1+cpu, PyG: 2.6.1 and scikit-learn: 1.2.1.  \\

\midrule
\multicolumn{2}{l}{\textbf{PrivGraph}} \\
Implementation & Adapted from the PGB implementation
\citep{liu2024pgb}; PrivGraph method due to \citet{Quan2023}. \\
Budget allocation & $\epsilon_1=\epsilon_2=\epsilon_3=\epsilon/3$; the final edge-count
and degree perturbations use $\epsilon_3$. \\

Community initialisation & $N=20$; nodes are initially assigned to consecutive blocks of at most
20 nodes, followed by random shuffling of community labels. \\

Initial community perturbation
& Off-diagonal counts receive Laplace noise with scale $1/\epsilon_1$;
diagonal counts use scale $2/\epsilon_1$, followed by non-negative post-processing. \\

Community refinement & Community detection uses resolution $t=1.0$ followed by the exponential-mechanism refinement with budget $\epsilon_2$. \\

Final edge counts
& Inter-community edge counts are perturbed with Laplace scale
$1/\epsilon_3$ and post-processed to be non-negative. \\

Within-community degrees
& Degrees are perturbed with Laplace scale $2/\epsilon_3$, post-processed,
and clipped to $[0,|C|-1]$. \\

Intra-community edges
& Generated using the implemented Chung--Lu-style construction from the
perturbed degree sequence. \\

Inter-community edges
& Sampled from the two endpoint communities according to the noisy
inter-community edge count and finally binarised. \\
Deviations/integration
& The implementation is wrapped for the TAPAS interface; node labels are
set to a default attribute value of zero. \\
Version / commit & PGB commit \texttt{076a910} \\

\bottomrule
\end{tabular}
\end{table}

\paragraph{Computational resources.}
Experiments were run on a compute server equipped with two Intel Xeon Gold 6254 CPUs (3.10\,GHz), 337\,GB RAM, and an NVIDIA RTX A6000 GPU. Graph generation and auditing were primarily CPU-based. The full EU-core experiment with $30$ target edges, six $\epsilon$ values, three auditors and both generators took approximately $10$ hours of wall-clock time over $15$ CPU cores.

\section{Technical Definitions}\label{apndx:defination}

This section provides additional definitions and technical details for the terminology and statistical procedures used in the main paper.

\subsection{Network Statistics}
\label{app:summaries}

We use the standard network summaries and notation of \cite{Barbour_Reinert_2026}. Given a graph $G(\mathcal{V}, \mathcal{E})$, with node set denoted by $\mathcal{V}$, edge set by $\mathcal{E}$, and with the adjacency matrix $A = \{a_{u,v}\}_{u,v \in \mathcal{V}}$.

\paragraph{Node degree.}

The degree $deg(v)$ of a node $v$ is the number of nodes $u$ that are neighbours of $v$:
\[deg(v) := \sum_{u \in \mathcal{V}, u\neq v} \mathbb{I}\{\{u,v\} \in \mathcal{E}\} = \sum_{u \in \mathcal{V}, u\neq v} a_{uv} = \sum_{u \in \mathcal{V}} a_{uv}\]
with the final equality following because, in a simple graph, $a_{vv} = 0$ for all $v \in \mathcal{V}$. The (immediate, closed) neighbourhood $\mathcal{N}(v)$ of a node $v$ consists of itself and the set of its neighbours, and has size $deg(v) + 1$.

For the GNN auditor, degree is computed within the edge-removed,
target-centred $k$-hop subgraph $H^{-e^\star(k)}_{u,v}$ and normalised by
the maximum degree in that subgraph:
\[
\tilde{d}(x) = \frac{d(x)}{\max_{z \in V(H^{-e^\star(k)}_{u,v})}d(z)}.
\]

\paragraph{Shortest Path Distance.} 
The (graph) distance $d(u,v)$ between two nodes  $u$ and $v$ is the length of a shortest path between them. We set $d(v,v)=0$ for every $v\in\mathcal{V}$ and $d(u,v)=\infty$ when no path exists between $u$ and $v$. Thus, adjacent
nodes are at distance one, and the immediate neighbourhood of $v$ can equivalently be written as
\[\mathcal{N}(v) := \{ u \in \mathcal{V} : d(u,v) \le 1 \}. \]
More generally, the $k$-hop neighbourhood of $v$ is
\[\mathcal{N}_k(v) := \{ u \in \mathcal{V} : d(u,v) \le k \}.\]
In particular, $\mathcal{N}_0(v) = \{v\}$ and $\mathcal{N}_1(v) = \mathcal{N}(v)$.

For the GNN auditor, shortest-path distances are computed in the edge-removed, target-centred subgraph $H^{-e^\star(k)}_{u,v}$. For a node $x$ in this subgraph, its distances from the target nodes $u$ and $v$ are normalised by the neighbourhood radius $k$:
\[\tilde{\delta}_u(x) = \frac{d(u,x)}{k}, \qquad \tilde{\delta}_v(x) = \frac{d(v,x)}{k}.  
\]

\paragraph{Local Clustering Coefficient.}

The local transitivity, or local clustering coefficient $C(v)$, of a node $v$ is the proportion of pairs of its neighbours that are themselves neighbours. A pair of neighbours that are adjacent forms a triangle with $v$, whereas a pair that is not adjacent forms a 2-star centred at $v$ but not a triangle. Thus, for $d(v)\geq 2$, the local clustering coefficient takes values between 0 and 1. Using the adjacency matrix, it can be written as
\[
C(v) = \frac{\sum_{u,w \in \mathcal{V},u \neq w} a_{uv}a_{wv}a_{wu}}{\sum_{u,w \in \mathcal{V},u \neq w} a_{uv}a_{wv}}.
\]

For the GNN auditor, the clustering coefficient is computed within the
edge-removed, target-centred $k$-hop subgraph. Thus, for each
$x\in V(H^{-e^\star(k)}_{u,v})$, the node feature is
\[
C_{H^{-e^\star(k)}_{u,v}}(x),
\]
the local clustering coefficient of $x$ computed using only the nodes and
edges contained in $H^{-e^\star(k)}_{u,v}$.

\paragraph{Common Neighbours.} For two distinct 
nodes $u$ and $v$, their number of common neighbours is
the number of nodes adjacent to both:
\[
\operatorname{CN}(u,v)
=
\left|
\left\{
x\in\mathcal{V}\setminus\{u,v\}:
\{u,x\}\in\mathcal{E},
\;
\{v,x\}\in\mathcal{E}
\right\}
\right|.
\]
For a simple undirected graph, equivalently, 
\[
\operatorname{CN}(u,v)
=
\sum_{x\in\mathcal{V}} a_{ux}a_{vx}.
\]

For the GNN auditor, common neighbours are computed within the edge-removed, target-centred $k$-hop subgraph $H^{-e^\star(k)}_{u,v}$:
\[
\operatorname{CN}_{H^{-e^\star(k)}_{u,v}}(u,v)
=
\left|
\Gamma_{H^{-e^\star(k)}_{u,v}}(u)
\cap
\Gamma_{H^{-e^\star(k)}_{u,v}}(v)
\right|,
\]
where $\Gamma_H(x)$ denotes the set of nodes adjacent to the node $x$ in graph $H$. This feature measures the extent to which the two target nodes share local structural context after the target edge itself has been removed.

\paragraph{Adamic--Adar.} 
The Adamic--Adar index \citep{ADAMIC2003211} assigns greater weight to common neighbours with smaller degree. For two nodes $u$ and $v$, it is defined as
\[
\operatorname{AA}_H(u,v)
=
\sum_{w\in\Gamma_H(u)\cap\Gamma_H(v)}
\frac{1}{\log d_H(w)}.
\]
where $d_H(w)$ denotes the degree of $w$ in $H$. Thus, common neighbours with smaller degree contribute more strongly, whereas high-degree common neighbours contribute less.

\subsection{Differentially Private Graph Generators} \label{sec:generators}
In this section, we provide additional details on the differentially private graph generators evaluated in our experiments: Randomised Response (RR) and PrivGraph. 

\paragraph{Randomised Response for Edge Differential Privacy.}

Randomised response (RR) \citep{karwa2014} independently perturbs graph edges to generate a synthetic graph. Let $X$ be the symmetric adjacency matrix of a simple graph on $n$ nodes, with $X_{uv}=\mathbb{I}\{u\sim v\}$. For an undirected graph, RR independently randomises the $n(n-1)/2$ possible edges to produce a released adjacency matrix $Y$.

For $p_{00},p_{11}\in[0,1]$, let
\[
p_{11}=\Pr(Y_{uv}=1\mid X_{uv}=1),\qquad
p_{00}=\Pr(Y_{uv}=0\mid X_{uv}=0).
\]
Thus, an edge is removed with probability $1-p_{11}$ and a non-edge is added with probability $1-p_{00}$. The resulting mechanism satisfies edge-DP with
\[
\epsilon =
\log \max \left\{
\frac{p_{00}}{1-p_{11}},
\frac{1-p_{11}}{p_{00}},
\frac{1-p_{00}}{p_{11}},
\frac{p_{11}}{1-p_{00}}
\right\}.
\]

In our experiments, we use symmetric randomised response, $p_{00}=p_{11}=1-\pi$, where $\pi$ is the probability of flipping the state of each dyad. The privacy parameter then reduces to
\[
\epsilon
=
\log\max\left\{
\frac{\pi}{1-\pi},
\frac{1-\pi}{\pi}
\right\}.
\]
For the regime $\pi\leq 1/2$ used here,
\[
\pi=\frac{1}{1+e^\epsilon},
\qquad
p_{00}=p_{11}=\frac{e^\epsilon}{1+e^\epsilon}.
\]

For example, $\epsilon=1$ gives $\pi\approx0.269$, while $\epsilon=0.1$
gives $\pi\approx0.475$. At $\pi=0.5$ ($\epsilon=0$), each released dyad is
independent of its original state, providing maximal privacy but destroying edge-level signal. Conversely, as $\pi\rightarrow0$, the released graph approaches the original graph while $\epsilon\rightarrow\infty$, illustrating the privacy--utility trade-off of RR. We assume $p_{00}$ and $p_{11}$ are public and therefore known to the auditor.

\paragraph{PrivGraph.}

PrivGraph \citep{Quan2023} is an edge-differentially private graph generator designed to preserve both community structure and local connectivity. Rather than perturbing individual edges independently, it first constructs a coarse partition of the nodes into communities and privately refines the partition using the exponential mechanism.

Given the partition, PrivGraph separately characterises connectivity within and between communities. For each community, it computes a noisy internal degree sequence, while inter-community structure is represented by noisy edge counts between pairs of communities. The privacy budget is distributed across these stages so that the overall procedure satisfies the required edge-level
DP guarantee.

A synthetic graph is then reconstructed from these noisy statistics. Edges within each community are generated from the perturbed degree information using a Chung--Lu (CL) model, while edges between communities are generated according to the corresponding noisy inter-community counts. Thus, PrivGraph preserves graph structure through a hierarchical representation: community membership captures coarse organisation, while degree information captures finer connectivity within communities.

This construction differs from edge-wise mechanisms such as RR because the released state of a particular edge is determined indirectly through noisy structural summaries rather than by independently perturbing that edge. Consequently, PrivGraph provides an interesting case for our audit: we test whether the membership of a target edge remains distinguishable either directly or through its surrounding local structural geometry after reconstruction.

\subsection{Interpreting differential privacy via a hypothesis test} \label{sec:hypdp}
The notion of $\epsilon-$ differential privacy can be expressed as a testing problem as follows; see, for example, \cite{wasserman_statistical_2009}. Suppose that an adversary observes a released graph \(H\) and tests
$ H_0 : H \sim \mathcal{M}(G')
   $ against $
    H_1 : H \sim \mathcal{M}(G),$ 
where \(G \sim G'\) are edge-neighbouring graphs. 

Recall that, for a measurable
rejection region \(A \subseteq \mathcal{G}\), we denote the true positive rate by 
$
    \operatorname{TPR}(A)
    = \Pr\!\left[\mathcal{M}(G) \in A\right],
$ and the {false positive} rate by $
    \operatorname{FPR}(A)
    = \Pr\!\left[\mathcal{M}(G') \in A\right].
$ 

If \(\mathcal{M}\) is \(\varepsilon\)-differentially private, then every
such test must satisfy
$$
    \operatorname{TPR}(A)
    \leq
    e^{\varepsilon}\operatorname{FPR}(A) 
\mbox{ and }
    1-\operatorname{FPR}(A)
    \leq
    e^{\varepsilon}\bigl(1-\operatorname{TPR}(A)\bigr) 
$$
Thus, differential privacy limits the ability of any hypothesis test to distinguish outputs generated from the two neighbouring graphs.
Conversely, a statistically supported test whose operating point is incompatible with these inequalities provides a witnessed lower bound on the privacy loss. 

\subsection{Confidence Bounds for Privacy Auditing} \label{sec:stat_bounds}

The empirical TPR and FPR are binomial proportions estimated from a finite audit set. To account for sampling uncertainty, we use exact one-sided Clopper--Pearson confidence bounds, following standard practice in empirical
DP auditing~\cite{nasr2021adversary,nasr2023tight}. 
\paragraph{Clopper--Pearson bounds.}
The Clopper--Pearson method \citep{clopper1934} constructs exact confidence bounds for a binomial proportion by inverting binomial tail probabilities. For a fixed decision threshold $\tau$, the number of positive decisions among the audit samples from each neighbouring graph distribution follows a binomial distribution. In general, if
\[
X\sim\operatorname{Binomial}(n,p)
\]
and $X=x$ successes are observed, for a one-sided error probability $\alpha$, the lower and upper confidence limits $L(x,n;\alpha)$
and $U(x,n;\alpha)$ satisfy
\[
\Pr_{p=L}(X\geq x)=\alpha,
\qquad
\Pr_{p=U}(X\leq x)=\alpha.
\]
Equivalently, using the relationship between binomial tails and the incomplete beta function, the same exact bounds can be expressed through Beta-distribution quantiles as
\[
L(x,n;\alpha)
=
F^{-1}_{\operatorname{Beta}(x,n-x+1)}(\alpha),
\]
and
\[
U(x,n;\alpha)
=
F^{-1}_{\operatorname{Beta}(x+1,n-x)}(1-\alpha),
\]
with the boundary conventions $L(0,n;\alpha)=0$ and $U(n,n;\alpha)=1$.

\medskip
For a fixed threshold $\tau$, let $L_{\mathrm{TPR}}$ denote a lower confidence bound on the TPR and $U_{\mathrm{FPR}}$ an upper confidence bound on the FPR so that 
\[
\operatorname{TPR}(\tau)=P_{\epsilon}(A_\tau\mid G), \quad \quad \operatorname{FPR}(\tau)=P_{\epsilon}(A_\tau\mid G').
\]
Let $n_1$ and $n_0$ denote the numbers of audit samples generated by the mechanism from $G$ and $G'$, respectively, and let $k_{\mathrm{TPR}}$ and $k_{\mathrm{FPR}}$ denote the corresponding numbers classified as originating from the distribution induced by $G$. We then obtain the one-sided bounds
\[
L_{\mathrm{TPR}}
=
L(k_{\mathrm{TPR}},n_1;\alpha'),
\qquad
U_{\mathrm{FPR}}
=
U(k_{\mathrm{FPR}},n_0;\alpha'),
\]
where $\alpha'$ denotes the per-bound error probability after the simultaneous-confidence correction described below.

Substituting these conservative bounds into the privacy-loss ratios yields the lower bound
$\epsilon_{\mathrm{LB}}$ in Eq.~\ref{eq:eps_lb}. Because $\epsilon_{\mathrm{LB}}$ is constructed from multiple estimated binomial probabilities, the required one-sided confidence statements must hold simultaneously. We therefore apply a Bonferroni correction, allocating the total error probability $\alpha=1-\gamma$ across the required bounds, where $\gamma$ is the desired simultaneous confidence level. By the union bound, all component bounds then hold jointly with probability at least $1-\alpha$, making $\epsilon_{\mathrm{LB}}$ a confidence-valid lower bound on the privacy loss witnessed by the auditor at the selected threshold.

\subsection{WL Structural Utility}
\label{app:wl_utility}

We measure structural utility using the normalised Weisfeiler--Lehman (WL)
graph kernel similarity; see \cite{shervashidze2011weisfeiler}. Let $K_{\mathrm{WL}}(G,H)$ denote the WL kernel of height $5$ between graphs $G$ and $H$. We use as base kernel the vertex-histogram kernel and constant initial node labels. For an original graph $G$ and a privatised graph $H$, we set 
\[
S_{\mathrm{WL}}(G,H)
=
\frac{
K_{\mathrm{WL}}(G,H)
}{
\sqrt{
K_{\mathrm{WL}}(G,G)
K_{\mathrm{WL}}(H,H)
}
}.
\]
Larger values indicate greater structural similarity between the original and privatised graphs.

For a target edge $e^\star$, let $G$ and $G'$ denote the two neighbouring
input graphs, and let
\[
H_1,\ldots,H_m\sim\mathcal{M}(G),
\qquad
H'_1,\ldots,H'_m\sim\mathcal{M}(G')
\]
be their corresponding privatised releases. We compute the mean WL similarity separately in each set of graphs,
\[
\overline S_G
=
\frac{1}{m}\sum_{i=1}^{m}S_{\mathrm{WL}}(G,H_i),
\qquad
\overline S_{G'}
=
\frac{1}{m}\sum_{i=1}^{m}S_{\mathrm{WL}}(G',H'_i),
\]
and report their equally weighted average,
\[
\overline S_{\mathrm{WL}}
=
\frac{1}{2}
\left(
\overline S_G+\overline S_{G'}
\right).
\]

Thus, each privatised graph is compared with the particular neighbouring input graph from which it was generated before averaging the mean utilities of the two samples. For each setting, we report the mean and standard deviation of $\overline S_{\mathrm{WL}}$ across the 30 sampled target edges. 

\section{Learned Structural Representations Using GNNs}

A more general alternative to manually specified structural features is to
learn representations directly from privatised graphs using a graph neural
network (GNN). Since the neighbouring input graphs $G$ and $G'$ differ only
in the target edge $e^\star=(u,v)$, we construct a target-centred
representation of the local structure surrounding $u$ and $v$.

Importantly, to isolate structural leakage beyond the observed state of the
target edge, we first remove $e^\star$ from every privatised graph $H$:
\[
H^{-e^\star}
=
\left(
V(H),
E(H)\setminus\{e^\star\}
\right) 
\]
and then construct the target-centred local subgraph 
$H^{-e^{\star}(k)}_{u,v} =
H^{-e^\star}
\left[
\mathcal{N}^{(k)}_{H^{-e^\star}}(u)
\cup
\mathcal{N}^{(k)}_{H^{-e^\star}}(v)
\right]$  as detailed in the next subsection. 
All subsequent neighbourhood extraction, node features, structural
statistics, and GNN message passing are computed on this edge-removed k-hop neighbourhood subgraph.
The representation-learning pipeline is therefore
\[
H
\longrightarrow
H^{-e^\star}
\longrightarrow
H^{-e^\star(k)}_{u,v}
\longrightarrow
\operatorname{GNN}_{\psi}
\longrightarrow
\{h_x\}
\longrightarrow
z_{\mathrm{GNN}}
\longrightarrow
f_{\omega}
\longrightarrow
\widehat{Y}.
\]

\subsection{Target-Centred Local Subgraph}

For a target node $u$, we denote its $k$-hop neighbourhood in the edge-removed
graph as
\[
\mathcal{N}^{(k)}_{H^{-e^\star}}(u)
=
\left\{
x\in V(H):
\operatorname{dist}_{H^{-e^\star}}(u,x)\leq k
\right\},
\]
and analogously for $v$. We take their union,
\[
V^{(k)}_{u,v}
=
\mathcal{N}^{(k)}_{H^{-e^\star}}(u)
\cup
\mathcal{N}^{(k)}_{H^{-e^\star}}(v),
\]
and construct the induced subgraph
\[
H^{-e^\star(k)}_{u,v}
=
H^{-e^\star}
\left[
V^{(k)}_{u,v}
\right].
\]
Thus, the target edge is absent not only from the explicit node features but also from the adjacency used to define the neighbourhood and perform GNN message passing. We use $k=2$ in all experiments.

\subsection{Initial Node Features}

Each node $x\in V^{(k)}_{u,v}$ is represented by the six-dimensional feature
vector
\[
\left[
\widetilde d(x),
\mathbb{I}\{x=u\},
\mathbb{I}\{x=v\},
\widetilde\delta_u(x),
\widetilde\delta_v(x),
C(x)
\right].
\]
All structural quantities are computed on
$H^{-e^\star(k)}_{u,v}$.
We abbreviate 
$
d(x)
=
d_{H^{-e^\star(k)}_{u,v}}(x).
$ 
The degree feature is normalised by the maximum degree within the extracted subgraph:
\[
\widetilde d(x)
=
\frac{d(x)}
{\max_{w\in V^{(k)}_{u,v}}d(w)}.
\]
If the maximum degree is zero, its denominator is set to one.

The distance features are shortest-path distances within the same local subgraph, scaled by the neighbourhood radius:
\[
\widetilde\delta_u(x)
=
\frac{
\operatorname{dist}_{H^{-e^\star(k)}_{u,v}}(u,x)
}{k},
\qquad
\widetilde\delta_v(x)
=
\frac{
\operatorname{dist}_{H^{-e^\star(k)}_{u,v}}(v,x)
}{k}.
\]
If $x$ is unreachable from one of the target nodes within the extracted subgraph, the corresponding distance is assigned $k+1$ before scaling. Hence, the unreachable-node encoding is $(k+1)/k$.

Finally,
\[
C(x)
=
C_{H^{-e^\star(k)}_{u,v}}(x)
\]
is the local clustering coefficient of $x$ computed within the extracted subgraph. The indicators $\mathbb{I}\{x=u\}$ and
$\mathbb{I}\{x=v\}$ identify the target endpoints without revealing whether the target edge appeared in the released graph.

\subsection{GNN-Based Representation Learning}

Given $H^{-e^\star(k)}_{u,v}$ and its node features, we apply a Graph Convolutional Network (GCN) to learn node representations. Let
\[
h_x^{(0)}= \left[
\widetilde d(x),
\mathbb{I}\{x=u\},
\mathbb{I}\{x=v\},
\widetilde\delta_u(x),
\widetilde\delta_v(x),
C(x)
\right].
\]
Each GCN layer aggregates information from neighbouring nodes in $H^{-e^\star(k)}_{u,v}$:
\[
h_x^{(\ell+1)}
=
\operatorname{GCN}^{(\ell)}
\left(
h_x^{(\ell)},
\left\{
h_w^{(\ell)}:
w\in
\Gamma_{H^{-e^\star(k)}_{u,v}}(x)
\right\}
\right).
\]
Our implementation uses three GCN layers with ReLU activations,
\[
h_x^{(1)}
=
\operatorname{ReLU}
\left(
\operatorname{GCN}^{(1)}(h_x^{(0)})
\right),
\]
\[
h_x^{(2)}
=
\operatorname{ReLU}
\left(
\operatorname{GCN}^{(2)}(h_x^{(1)})
\right),
\qquad
h_x^{(3)}
=
\operatorname{ReLU}
\left(
\operatorname{GCN}^{(3)}(h_x^{(2)})
\right),
\]
with final node representation $h_x=h_x^{(3)}$. Since the target edge is removed before constructing the GNN input, message passing can exploit only its surrounding local structural geometry.

\subsubsection{Target-Pair and Subgraph Representation}

Since the target edge is undirected, its representation should be invariant to exchanging $u$ and $v$. We therefore construct the symmetric endpoint representation
\[
z_{\mathrm{pair}}
=
\left[
h_u+h_v
\,\Vert\,
|h_u-h_v|
\,\Vert\,
h_u\odot h_v
\right],
\]
where $\Vert$ denotes concatenation and $\odot$ element-wise multiplication. This satisfies
\[
z_{\mathrm{pair}}(u,v)
=
z_{\mathrm{pair}}(v,u).
\]

To capture information distributed throughout the target neighbourhood, we also compute mean and element-wise maximum pooling:
\[
g_{\mathrm{mean}}
=
\frac{1}{|V^{(k)}_{u,v}|}
\sum_{x\in V^{(k)}_{u,v}}h_x,
\qquad
g_{\mathrm{max}}
=
\operatorname*{max}_{x\in V^{(k)}_{u,v}}h_x.
\]

Finally, we compute the common-neighbour count of the target pair on the edge-removed local subgraph,
\[
\operatorname{CN}(u,v)
=
\left|
\mathcal{N}_{H^{-e^\star(k)}_{u,v}}(u)
\cap
\mathcal{N}_{H^{-e^\star(k)}_{u,v}}(v)
\right|.
\]
This is a target-pair feature and is separate from the six-dimensional node feature vector above. The final GNN representation is
\[
z_{\mathrm{GNN}}
=
\left[
h_u+h_v
\,\Vert\,
|h_u-h_v|
\,\Vert\,
h_u\odot h_v
\,\Vert\,
g_{\mathrm{mean}}
\,\Vert\,
g_{\mathrm{max}}
\,\Vert\,
\operatorname{CN}(u,v)
\right].
\]

\subsubsection{Classification and Training}

For  privatised graphs from the two neighbouring graph distributions,
\[
H_i
\overset{\mathrm{i.i.d.}}{\sim}
P_{\epsilon}(\cdot\mid G),
\qquad
Y_i=1,
\]
and
\[
H'_i
\overset{\mathrm{i.i.d.}}{\sim}
P_{\epsilon}(\cdot\mid G'),
\qquad
Y'_i=0.
\]
The representation $z_{\mathrm{GNN}}$ is passed to a multilayer perceptron $f_\omega$ to produce the attack score
\[
s_H
=
f_\omega(z_{\mathrm{GNN}}),
\qquad
\widehat p_H
=
\sigma(s_H),
\]
where $\widehat p_H$ is the predicted probability of $Y=1$. The GNN parameters $\psi$ and classifier parameters $\omega$ are
trained jointly using binary cross-entropy.

The resulting auditor therefore receives the identities of the target nodes but never their released edge state: $e^\star$ is removed before neighbourhood construction, feature computation, and message passing. Consequently, successful distinguishability provides evidence that information about the target edge persists in the surrounding local structural representation rather than being obtained by directly observing the released target edge.


\end{document}